\documentclass{article}

\usepackage[utf8]{inputenc} 
\usepackage[T2A,T1]{fontenc} 
\usepackage[russian,english]{babel}

\usepackage[scaled=0.9]{PTMono}

\usepackage{enumitem}

\usepackage{graphicx}

\usepackage{stfloats}

\newcommand{\rustt}[1]{%
  \begingroup%
  \fontfamily{ptm}\selectfont%
  \foreignlanguage{russian}{#1}%
  \endgroup%
}

\DeclareUnicodeCharacter{20BD}{\textruble}

\providecommand{\textruble}{%
  \makebox[0pt][l]{\hspace{0.08em}--}P%
}

\usepackage{tabularx} 
\usepackage{booktabs} 
\usepackage{ragged2e} 

\usepackage{xltabular}

\usepackage{microtype}
\usepackage{graphicx}
\usepackage{subcaption}
\usepackage{booktabs} 

\usepackage{hyperref}

\usepackage[preprint]{fancypaper}

\usepackage{amsmath}
\usepackage{amssymb}
\usepackage{mathtools}
\usepackage{amsthm}

\usepackage[capitalize,noabbrev]{cleveref}

\theoremstyle{plain}

\theoremstyle{definition}

\theoremstyle{remark}

\usepackage[textsize=tiny]{todonotes}

\icmltitlerunning{asr\_eval: algorithms and tools for multi-reference and streaming speech recognition evaluation}

\newcolumntype{L}{>{\raggedright\arraybackslash}X}

\begin{document}

\twocolumn[
  \icmltitle{asr\_eval: algorithms and tools for multi-reference and streaming speech recognition evaluation}



  \icmlsetsymbol{equal}{*}

  \begin{icmlauthorlist}
    \icmlauthor{Oleg Sedukhin}{sib}
    \icmlauthor{Andrey Kostin}{sib}
  \end{icmlauthorlist}

  \icmlaffiliation{sib}{Siberian Neuronets LLC, Novosibirsk, Russia}

  \icmlcorrespondingauthor{Oleg Sedukhin}{sedol1339@gmail.com}

  \icmlkeywords{Machine Learning, asr, alignment, streaming, longform, evaluation}

  \vskip 0.3in
]



\printAffiliationsAndNotice{}  

\begin{abstract}
  We propose several improvements to the speech recognition evaluation. First, we propose a string alignment algorithm that supports both multi-reference labeling, arbitrary-length insertions and better word alignment. This is especially useful for non-Latin languages, those with rich word formation, to label cluttered or longform speech. Secondly, we collect a novel test set DiverseSpeech-Ru of longform in-the-wild Russian speech with careful multivariant labeling. We also perform multivariant relabeling of popular Russian tests set and study fine-tuning dynamics on its corresponding train set. We demostrate that the model often adopts to dataset-specific labeling, causing an illusion of metric improvement. Based on the improved word alignment, we develop tools to evaluate streaming speech recognition and to align multiple transcriptions to compare them visually. Additionally, we provide uniform wrappers for many offline and streaming speech recognition models. Our code will be made publicly available.
\end{abstract}

\section{Introduction} \label{introduction}

\begin{figure*}[h!]
    \centering
    \fbox{\includegraphics[width=\linewidth]{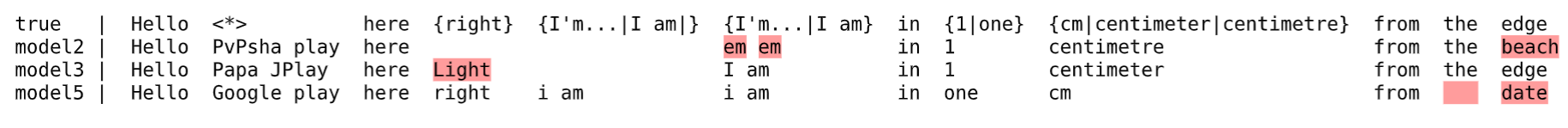}}
    \caption{Multiple transcription alignment.}
    \label{fig:alignment}
\end{figure*}

Speech recognition evaluation is complicated by multiple possible
spellings, cluttered and overlapping speech. This is typically solved by
text normalization, but it does not cover all cases. Table \ref{use-cases-short} summarizes
multiple kinds of complexities that arise when annotating speech, and
the possbility to solve them by text normalization (see the extended version with discussions in Appendix \ref{app:extended_table}). Also, a good
normalization model may not exist for every language. In table \ref{use-cases-short}, we
employ a syntax that uses braces to list multiple options, and a
wildcard symbol for unclear speech.

In this paper, we propose \textbf{MWER}\footnotemark, a string alignment
algorithm that can parse this syntax directly and align a prediction
against a transcription with multi-reference blocks, while aligning
wilcard symbol with arbitrary word sequence, possibly empty. Also, MWER
modifies a traditional scoring function to improve word-to-word
alignment. Finally, when we use the alignment to calculate Word Error
Rate (WER), we use a relaxed insertion penalty to make our metrics more
stable against oscillatory hallucinations. In our code implementation,
each aspect is customizable and can be turned off.

We also release \textit{\textbf{asr\_eval}}, ASR evaluation library for
Python. It provides tools to (i) extend annotation syntax via MWER, (ii)
get insights into model performance via dashboard and streaming plots,
(iii) use a collection of ASR building blocks to
avoid code duplication, and model and datasets in a unified format. It includes:

\footnotetext{Acronym for \textbf{M}ulti-reference \textbf{W}ildcard and \textbf{E}nhanced alignment with \textbf{R}elaxed insertion penalty}

\begin{enumerate}[topsep=0pt]
\item
  The whole evaluation pipeline with tokenizing, preprocessing, aligling
  using the proposed MWER algorithm, WER/CER calcuiation and
  fine-grained error analysis, with extensive documentation and intermediate data structures for practioners with custom needs.
\item
  Multiple word-to-word alignments to compare several models, and an
  interactive dashboard to visualize them. Error highlighting feature
  can be used to rapidly validate datasets and correct annotation errors
  in them, as we show in section \ref{multiple-alignment-and-dashboard}.
\item
  A collection of datasets and model wrappers in standardized format,
  and a framework to run inference and store the results, with dashboard
  integration. It allows for custom text normalizers, tokenizers and
  applying noise overlays.
\item
  Building blocks, such as longform VAD wrapper for any ASR
  model + voice activity detector, and longform CTC wrapper for any
  CTC model, to support long audios. This allows to quickly combine
  different components and evaluate them, including for ablation
  studies.
\item
  Streaming inference and evaluation tools for research and production.
  This includes a flexible base class to process several audio streams
  simultaneously, a storage mechanism for the input-output history and
  various evaluation diagrams. Specifically, we propose (i) \emph{time remapping} for faster evaluation, (ii) a
  \emph{streaming alignment diagram} for a single sample, and (iii) a
  \emph{streaming histogram} for aggregating.
\end{enumerate}

We also relase \textbf{DiverseSpeech-Ru}, a Russian longform
Youtube-sourced dataset. We propose an annotation guideline for
multi-reference annotation with wilcard insertions that MWER algorithm
can handle. This is especially useful for non-Latin languages, languages
with rich word formation, to label cluttered or longform speech. We make
use of our multiple alignment feature in dashboard to quickly validate
the dataset: if many baseline models made a ``mistake'' in some place,
one needs to check the annotation in it.

Finally, we conduct exepriments to check if the MWER features matter in
practice. We relabel an existing Russian dataset using careful
multi-reference and wildcard syntax, and compare the re-labeleing
approach with the text normalization approach, studying fine-tuning
dynamics. We observe that both versions show different fine-tuning
dynamics. Since the text normalization approach is apriori approximate,
and the multi-reference re-labeling appoach is more strict, their
different fine-tuning dynamics suggests that the model adopts to
dataset-specific labeling, causing an illusion of metric improvement
even with normalization. This is crucial in research, when we can
confuse the actual efficient fine-tuning method with the described
artifact.

\begin{table*}[t] 
    \centering
    \caption{\textbf{Multi-reference and wildcard syntax use cases.} We use syntax \texttt{\{A|B|C\}} to enumerate multiple equally acceptable options, \texttt{\{A\}} for optional text blocks, \texttt{\{A|~B\}} for the case where B is acceptable but contains minor typical spelling errors. \texttt{<*>} is a wildcard symbol used where we cannot reliably enumerate all possible choices. Our MWER algorithm can use the provided syntax directly to evaluate WER/CER, aligning \texttt{<*>} with arbitrary sequence without penalty. \textit{The extended version of this table with more discussions can be found in Appendix \ref{app:extended_table}, and complex cases are described in Appendix \ref{app:complex_cases}.}}
    \label{use-cases-short}
    
    \begin{tabularx}{\textwidth}{l L L} 
        \toprule
        \textbf{Use case} & \textbf{Examples} & \textbf{Can text normalization handle this?} \\
        \midrule
        Numerals, currency and units of measurement
        & 
        En: \rustt{\{Fourth|4|4th|4-th\}} \newline En: \rustt{\{10|ten\} \{mm|millimeters\}}
        & 
        Yes, but may be prone to errors.
        \\
        \midrule
        Poorly heard inflections
        & 
        En: \rustt{the player's own \{fantasy|fantasies\}}. \newline Ru: \rustt{частичку \{фантазий|фантазии\} игрока}.
        & 
        Typically no.
        \\
        \midrule
        Wrongly spelled inflections
        & 
        Ru: \rustt{Вот наши \{шампура|шампуры\}}.
        & 
        Typically no.
        \\
        \midrule
        Repetitions
        & 
        En: \rustt{And \{the first we need...\} the first we need is a plan.} \newline Ru: \rustt{И \{первое, что нам нужно\} первое, что нам нужно это план.}
        & 
        Typically no.
        \\
        \midrule
        Transliterations, possibly inflected ones
        & 
        Ru: \rustt{в \{Facebook|Facebook-е|Фейсбуке\}}
        & 
        Sometimes.
        \\
        \midrule
        Poorly heard speech, hesitations.
        & 
        \rustt{<*>}
        & 
        No.
        \\
        \midrule
        Filler words
        & 
        En: \rustt{\{well\}} \newline Ru: \rustt{\{угу|ага|\} , конечно.}
        & 
        Partly (see Appendix \ref{app:extended_table} for discussions).
        \\
        \midrule
        Minor syntax errors
        & 
        Ru: \rustt{\{неудивительно|~не удивительно\}}
        & 
        Typically no.
        \\
        \bottomrule
    \end{tabularx}
    
\end{table*}

\section{Related work} \label{related-work}

In automatic speech recognition, both WER (word error rate) and CER
(character error rate) metrics are based on the optimal sequence
alignment. The core algorithm to align two sequences is a
Needleman-Wunsh algorithm \citep{NEEDLEMAN1970443} based on filling a matrix via dynamic
programming that produces an optimal alignment between sequences. The
Wagner--Fischer algorithm \citep{wagner1974string}, when used in ASR, is a special instance of the general
Needleman--Wunsch algorithm that calculates only the edit distance.

In the past years, \citep{fiscus2006multiple} extended a string alignment aligorithm
by represent a ground truth as a directed acyclic graph, thus allowing
to handle multiple ground truths, similarly to our work. They focus
solely on multi-speaker evaluation challenges, and thus do not handle
arbitrary insertions and do not improve a scoring function for better
alignment. \citep{von2023word} also focus on multi-speaker evaluation and
extend the Wagner--Fischer algorithm to support a compositional ground
truth of muitple-speaker utterances. \citep{karita2023lenient} note the importance
of orthographic variation for Japanese language: different spellings may
be equally accepted. They train a model to automatically generate
multi-reference transcriptions from single-reference ones. Later, \citep{kaji2023lattice}
utilized multivariance to enumerate all possible romanized forms of
Japanse words.

Similar problems arise when aligning DNA sequences in biology.
Elastic-degenerate (ED) strings were proposed to align a genome against
a pangenome (a set of closely-related DNA sequences). \citep{gabory2025elastic}
proposed al algorithm to align two ED-strings in a similar way as
\citep{fiscus2006multiple} did in multi-speaker speech recognition. Their formulation
allows empty option in a multivariant block, but not arbibrary length
insertions, since ED-string is a sequence of finite sets of strings. In
biology multiple sequence alighment (MSA) is commonly used to align
several DNA sequences with each other. A similar operation was proposed
to align multiple ASR hypotheses from different models, known as ROVER
system \citep{fiscus1997post}.

\citep{mcnamara2024style} make multi-reference annotations for two ASR evaluation
sets of spontaneous long-form speech. They annotated each audio twice:
with verbatim (including filler words etc.) and and less verbatim
transcriptions. Notably, they show that such a double annotation results
in WER lower than any kind or normalization they tried, showing that
normalization does not cover all cases. This is in line with ours work,
however, we are interested if using multivaraince instead of
normalization may affect \emph{model ranking} (e.g. selecting the best
fine-tuning checkpoint) and not just the absolute WER values. We
describe our experiments in more details in section \ref{fine-tuning-dynamics}. Thus, incorporating
multivariance gains interest in the research commnunity. As far as we
know, still none of the previous works consider arbitrary-length
insertions (that may be useful for poorly heard speech) or improving
word alignment.

\section{Transcription format} \label{transcription-format}

We propose an annotation format, guidelines and alignment algorithm that
supports both multiple references and arbitrary-length (wildcard)
insertions, and improve scoring function to obtain better word
alignment. We release a code base to support further research.

\subsection{Multi-reference blocks} \label{multi-reference-blocks}

A multi-reference block is a ground truth transcription section with
multiple correct options, each with zero or more elements. An element is
typically a word, or a character if we perform character-level
alignment. Our syntax extends regular (single variant) transcriptions.
An annotator writes a multi-reference blocks in curly brackets,
separating options with a vertical line (pipe) symbol. Each option may
contain multiple words which are tokenized in the same way as regular
text (our tokenization rule is customizable). This is useful in several
scenarios not always handled by normalization, see examples in Table \ref{use-cases-short}.
To simplify adding optional words, we use an additional parsing rule: if
a block contains only a single option, an empty option is added
automatically during parsing. That is, \rustt{\{well\}} is equivalent as
\rustt{\{well\textbar\}}, but \rustt{\{one|1\}} is \emph{not} equivalent
to \rustt{\{one|1|\}}.

When annotating DiverseSpeech-Ru (Section \ref{novel-datasets}) our general rule was to
include options acceptable in the case of a human transcriber who is
generally literate, but not necessarily perfectly literate. There are
also rare and complex cases we show in discuss in Appendix \ref{app:complex_cases}.

Should we add options with minor spelling errors? This is up to
practioners, but we argue that sometimes we should. In some languages
many rules are too complex, and some words are often spelled
incorrectly. We extend our syntax to support prepending tilde
``\textasciitilde{}'' to options with minor syntax errors. This symbol
is removed during parsing and is turned into a flag. This allows to
evaluate in two scenarios: lexically strict and permissive.

Does it complicate the annotation process? We believe that multivariance
usually makes things easier, because most cases where it is required are
obvious to annotator, and you no more need to choose between equally
acceptable options (and incorporate your own selection biases into a
dataset). However, if you go further and try to make your annotation
perfect, you find complex cases where multi-reference annotation is
controversial (Appendix \ref{app:complex_cases}), but the percentage of such cases is small.

\subsection{Wilcard insertions} \label{wilcard-insertions}

We use a special wilcard symbol \rustt{<*>} in a ground
truth transcription that matches with any word sequence, including empty
sequence. This allows to skip annotating some utterances that contain
poorly heard or cluttered speech by permitting any transcription for
them. This is especially useful in annotating longform speech, because
almost every longform speech sample would contain at least one section
we're not sure about. If an annotator instead tries to annotate every
poorly head speech, they would likely introduce their own biases that
may favor some models and especially checkpoints fine-tuned on a split
of the same dataset.

However, using \rustt{<*>} is discouraged when we
actually know what was said, despite the speech being poorly heard. This
is the case when (i) the speaker is the annotator, (ii) we have a text
that was read, (iii) we add artificial background noise or simulate bad
microphone. Also note that extensive using
\rustt{<*>} for poorly heard utterances largely
lowers absolute WER values.

\section{Evaluation improvements} \label{evaluation-improvements}

\subsection{Transcription alignment} \label{transcription-alignment}

We provide MWER algorithm overview here and describe in more details in
Appendix \ref{app:appendix_algorithm}.

First, it can be seen that the optimal alignment algorithm can be easily
solved via recursion: we can take the first element from both sequences
and thus reduce a problem of alignling sequences of length (A, B) to
aligning sequences of length (A-1, B), (A-1, B-1) and (A, B-1). Thus,
the aligning requires 3 recursive calls. The same idea can be
implemented without recursive calls, by filling a matrix of size (A, B):
each cell can be filled based on 3 other neighbour cells, just like in
recursion. This idea is the basis of the Needleman-Wunsh algorithm.

\begin{enumerate}[topsep=0pt]
\item
  We extend the algorithm to multi-reference case similarly to \citep{fiscus2006multiple}.
  All the elements are first enumerated: for example, the
  transcription \rustt{"\{A|B\} \{C\} D"} is enumerated as A, B, C, D,
  forming 4 matrix rows. In contrast with the Needleman-Wunsh algorithm,
  we allow "jumps" between non-neighbour rows based on the element
  connectivity: for example, we can jump from A to D, because the block
  \rustt{\{C\}} is optional.
\item
  We support wilcard insertion symbol \rustt{<*>}
  that matches with any word sequence. In the Needleman-Wunsh algorithm,
  horizontal transition in the matrix (to the next predicted token,
  keeping the ground truth token the same) increases the error counter,
  because this is actually an insertion (extra preidcted token). In our
  case, if the ground truth token is \rustt{<*>},
  horizontal transitions do not increase the error counter.
\item
  In the Needleman-Wunsh algorithm, the alignment score (calculated for
  each matrix transition) is an integer. In our case, the score is a
  tuple of 3 numbers compared lexicographically. The first number is the
  word error count, as usual. The second number is the number of total
  correct matches, and the last number is a sum of all character errors
  for each match. For example, matching "hello" with "hey" is a
  replacement treated as 1 word error and 3 character errors. Note that
  this does \emph{not} calculate CER for the whole text, but is useful
  to align words better (see Section \ref{improved-alignment-scoring}).
\end{enumerate}

\subsection{Improved alignment scoring} \label{improved-alignment-scoring}

Consired aligning a prediction "multivariant" against the ground truth
"multivariate though". We can align "multivariant" against the first
ground truth word and let the second be deletion, or vice versa. Both
alignments are equally optimal in sence of WER: they both give 1
replacement and 1 deletion. But obviously, one of them is better. This
makes sense in streaming evaluation where determining a ground truth
word for a predicted word is a required step to evaluate latency (even
if this word is wrongly transcribed). Also, this visually improves the
alignment to compare multiple offline models on a single sample. A valid
alternative is to use character-level alignment, but this is difficult
to combine with using WER (and not CER) as a metric or interest.

The Needleman-Wunsh algorithm compare scores of multiple possible
alignments, where score is a total number of errors. The goal is to find
the alignment with minimum score. However, as in the example above,
there are many alignments with minumum score, and the choice among them
is arbitrary.

We modify the objective to be a tuple of scores compared
lexicographically. The first element in tuple is a number of errors
(word errors if we align on a word level). The next elements are
customizable and matter only if the number of errors is the same. Thus,
we keep the main goal to minimize number of errors, but also optimize
additional metrics. Currently, in our score tuple we use two additional
metrics we describe in Appendix \ref{app:appendix_algorithm}, one of them is CER-based, another is
the number of correct matches.

Typically, this gives visually better alignment. However, we do not
prove that our concrete list of scores gives the best word-to-word
alignment, because it's not clear what "best" means. Though we
believe this question is important to calculate streaming latency or do
a fine-grained error analysis, and our mechanism for adding additional
scores is rather general and can be experimented with, using our code
base, in future work.

\subsection{Relaxed insertion penalty} \label{relaxed-insertion-penalty}

Generative speech recognition models wuch as Whisper \citep{radford2023robust} are sometimes prone
to oscillatory hallucinations \citep{guerreiro2023hallucinations}, where the model generates the same
token, word of phrase until the generation limit. In traditional WER/CER
formulation, each hallucination hugely affacts metrics, making the
metric unstable. Mathematically speaking, the metric distribution become
heavy-tailed because of rare but long oscillatory insertions. This
hinders evaluation if we have limited data. Simply, one checkpoint makes
a \emph{random} hallucination in one sample, which lowers its average
metrics hugely.

We tackle this by treating more than 4 subsequent insertions as exactly
4. This should also align better with human evaluation, because the same
word repeated 100 times does not hinder text understanding so much as
100 word errors in different places.

However, the problem is that "hallucinations" are not always
oscillatory. If the model hallucinates on non-speech noise, this may
result in long insertions that do not repeat the same word. This hinder
text understanding more. Thus, we plan to experiment with another
approach to turn a long insertion into set, to evaluate the number of
different words in it.

\section{asr\_eval evaluation tools} \label{asr-eval-evaluation-tools}

In this section we describe parts of the proposed \textit{asr\_eval} library,
besides the alignnment algorithm.

\subsection{Multiple alignment and dashboard} \label{multiple-alignment-and-dashboard}

We develop a dashboard to visualize multiple alignments between model
predictions and the ground truth, making use of our tweaks to improve
word-to-word alignment (Figure \ref{fig:alignment}) and develop
a web interface (Figure \ref{fig:dashboard_screenshot}). This turns out to be useful when
preparing high-quality evaluation sets. We further describe our case
study.

We first annotated 92 recordings 2-3 minutes long with multi-reference
labeling. This data was annotated by one annotator and verified by
another annotator. Then we obtained a set of predictions for each sample
with GigaAM, several Whisper configurations, Voxtral etc. We align and
visualize the predictions in a dashboard with highlighting errors. As a
result, we spot and fix many annotation errors. Some of them were plain
misspellings. Another errors are more subtle: they arise when the
annotator focuses on a single option to transcribe and ignores another
possible options. However, some models transcribe differently. By
analyzing multiple alignments the annotator concludes that the
highlighted "errors" are caused by missing options in the annotation.
We also analyzed many other existing datasets and observe similar cases,
including misspellings.

Thus, our instrument helps to quickly analyze and fix existing datasets,
either multi-reference or not. Using a dashboard, we simultaneously
evaluate both models and the annotation, including annotation verbosity,
misspellings, missing words etc.

\subsection{Wrappers and building blocks} \label{wrappers-and-building-blocks}

We wrap multiple models into a unified programming interface, including Whisper \citep{radford2023robust},
Voxtral \citep{liu2025voxtral}, wav2vec2 \citep{baevski2020wav2vec}, \citep{puvvada2024less},
Qwen2-Audio \citep{chu2024qwen2}, GigaAM \citep{kutsakov2025gigaam} and more, into one of the 4 interfaces:
\verb|Transcriber|, \verb|TimedTranscriber|, \verb|CTC| and \verb|StreamingASR|.

We also gather many existing datasets (mostly Russian
currently). For each dataset, we apply tools that detect possible
train-test sample overlap and estimate speaker count and train-test
speaker overlap. For each dataset, we report our analysis in
documentation and fix sample overlaps if found.

We provide several building blocks to assemble transcribers. WWe believe
this is important for ablation studies and disentangling influence of
components, and to avoid duplicating logic. The main bulding blocks:

\begin{enumerate}[topsep=0pt]
\item
  \verb|CTCDecoderWithLM| can wrap around any CTC model to decode with
  KenLM using \textit{pyctcdecode} \citep{pyctcdecode2022}.
\item
  \verb|LongformVAD| combines any shortform transciber and voice
  activity detector to process long audios.
\item
  \verb|LongformCTC| wraps around any CTC model to chunk audio
  uniformly and merge logit matrices. It spoorts several merging
  algorithms, and in prelimilary studies, we observe that the best
  merging algorithm may be model-dependent.
\item
  \verb|ContextualLongformVAD| wraps around a
  \verb|ContextualTranscriber|, which is typically an audio-LLM that
  can accept the previous transcription and the current audio, similarly
  to Whisper.
\end{enumerate}

\subsection{Streaming evaluation} \label{streaming-evaluation}

Streaming speech recognition system can be evaluated by a variety of
parameters: (i) final WER/CER, (ii) latency induced by calculations, (iii)
latency induced by accumulating context, where the model refuses to emit
a word until accumulates enough audio frames, even with infinite compute
speed, (iv) the dependency between average WER and time delta, if the
system is able to correct previous transcription as more audio context
comes, (v) distributions, spikes and edge cases of the parameters
described above.

To tackle this complexity, we take several steps. First, to unify
evaluation, we define a general input-output interface: the system
accepts audio chunks labeled by recording ID via input buffer, and emits
strings labeled by recording ID and part ID. When a new string is
emitted, if the part ID matches that for one of the previous parts, it
replaces that part, otherwise the new string appended to the end of the
text. This allows a system to correct previously emitted words or
sentences, if needed. For the described interface, we develop a wrapper
with input and output buffers, which makes it easy to wrap any streaming
ASR into our interface.

\begin{figure*}[h!]
    \centering
    \includegraphics[width=\linewidth]{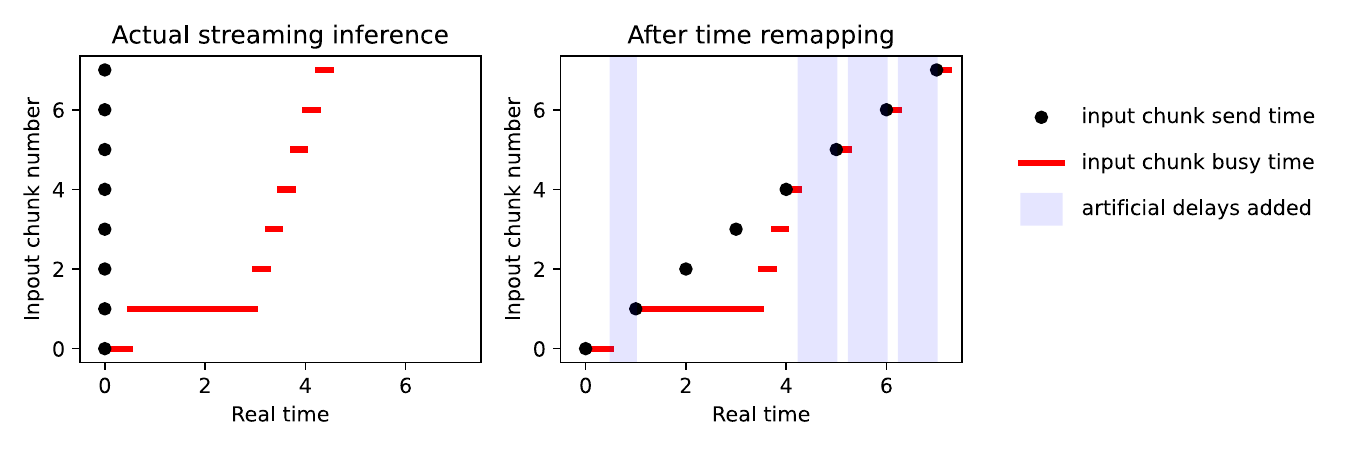}
    \caption{\textbf{Time remapping in streaming evaluation.} X axis is a real
        time, and Y axis is a chunk number (the audio consists of 5 chunks). For
        each chunk, we mark send time (black dot) and time span when the system
        was busy processing this chunk (red). Left: all chunks were send at
        once. Right: all chunks were sent were sent evenly in real time. Blue
        denote idle time we want to eliminate. Time remapping takes chunk
        history from the process on the left and alters the recorded timestamps
        to mimic process on the right. Thus we get the real time simulations
        faster than real time.}
    \label{fig:remapping}
\end{figure*}

To simulate real-time CPU/GPU loading (and thus a realistic latency) we
need to send audio in real time. However, this leads to time spans when
both the sender and the system waits, because the system had processed
all the previous chunks, but the time to send the next chunk has not yet
come. To speed up simulated evaluations, we may wish to eliminate these
spans. To this end, we develop an algorithm called \emph{time
remapping} (Figure \ref{fig:remapping}): we send all the input chunks at once, process them, and
during evaluation we add artificial time spans to accurately simulate
real-time sending. This process can be understood from the fig.~2. This
is, however, limited to single-threaded systems.

\begin{figure}[h!]
    \centering
    \includegraphics[trim=0.5cm 0cm 0.5cm 0.5cm, clip, width=\linewidth]{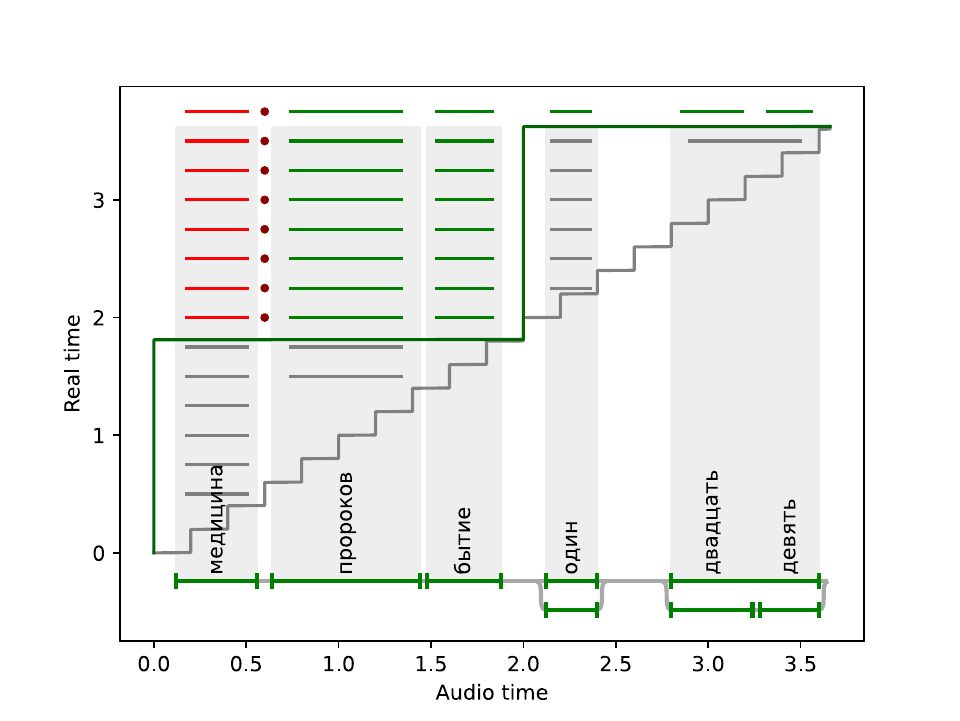}
    \caption{\textbf{A streaming evaluation diagram.} Each row is a partial
        alignment, where correctly transcribed words are show in green, wrongly
        transcribed words are shown in red, insertions (extra words) are shown
        as red dots (their timestamps are fictitious), deletions (missed words)
        and words not yet transcribed are shown in gray.}
    \label{fig:streaming_alignment}
\end{figure}

In view of the diversity of all the streaming characteristics, we
believe that diagramming is a good way to analyze a streaming system. We
start evaluation from a history of input and ouptut chunks and a ground
truth transcription. First, we perform forced alignment with CTC model
to obtain word-level ground truth timestamps \footnote{CTC ovjective by design does not force a model to predict timestamps
accurately, but empirically timestamps are pretty accurate in most
cases. However, there may be exceptions, so care should be taken.}. Then we take evenly
growing moments T\_1, \ldots, T\_N in real time to evaluate at. For each
time moment T\_i, we take all the output chunks emitted before, and
merge them into a partial transcription at T\_i. We also consider the
audio span sent by T\_i and all its words, according to the word-level
timestamps. We then align the partial transcription against the partial
ground truth. If the T\_i is in mid-word, we consider this word
optional, making use of multi-reference syntax. As a result, we obtain a
partial alignment, consisting of "correct", "replacements",
"insertions" and "deletions", as usual. Additionally, we extract a
tail of words marked as "deletions" and mark them "not yet
transcribed". The result is shown in Figure \ref{fig:streaming_alignment}. Here we use the improved
word-to-word alignment in MWER algorithm, otherwise our word matches
could be wrong - for example, replacements and deletions could be
falsely swapped, as in the example in section \ref{improved-alignment-scoring}.

The proposed streaming evaluation diagram provides insights into the
model behaviour. It allows to spot lags at some chunks, missing words
and more problems. \textit{asr\_eval} provides each diagram as a plot and
as a data structure, enabling any custom aggregation into numerical
metrics.

\begin{figure}[h!]
    \centering
    \includegraphics[width=\linewidth]{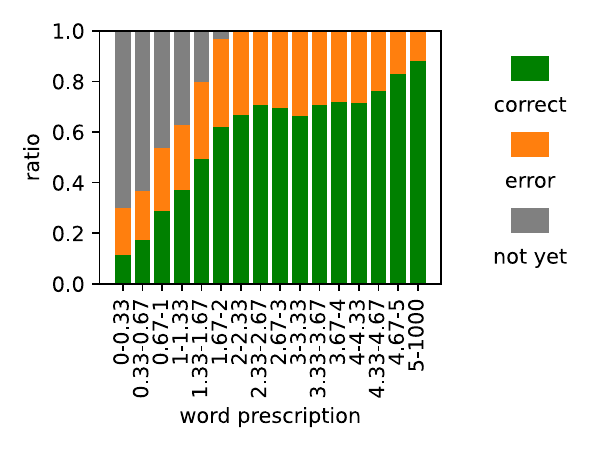}
    \caption{\textbf{A streaming quality historgram.} See Section \ref{streaming-evaluation} for description.}
    \label{fig:streaming_histogram}
\end{figure}

The streaming evaluation diagram visualizes a single sample. To
aggregate metrics for multiple samples, we propose a \emph{streaming
histogram}. We first take each word for each partial alignment and for
each sample. For each word, wa calculate \emph{prescription} ("how long
ago was the word spoken") - the difference between the word timing
(center time) and the length of the audio sent at time of the current
partial alignment. Also, we categorize each word into one of 3 types:
(i) correct, (ii) not yet transcribed, (iii) errors - replacement,
insertion or deletion (excluding not yet transcribed tail). So, for each
word we define prescription and one of 3 types. This is enough to make a
histogram (Figure \ref{fig:streaming_histogram}). We divide prescription into bins (X axis), and for
each bin we count the ratio of all types: correct (green), errors
(orange) and not yet transcribed (gray). This provides insights into
system's latency and quality. For example, a system that runs a model on
the whole received audio 2 times a second, will exhibit high error ratio
at low prescription, because the last word is usually patrially cut off.
In contrast, a system that accumulates context to transcribe will
exhibit high "not yet transcribed" ratio but lower error ratio at low
prescription. Also, if a system incudes additional slower model to
finalize transcription, the error ratio will be lower at high
prescription. These are valuable insights into model characteristics.

\section{Novel datasets} \label{novel-datasets}

We publicly release two Russian test sets for speech recognition:

\begin{enumerate}[topsep=0pt]
\item
  A longform speech dataset with a thorough multi-reference labeling,
  containing 3.5 hours of speech in form of 2-3 munute long audios. We
  take YODAS dataset \citep{li2023yodas} as a source (a "YODAS2" version with full audios)
  and use correction based on multiple alignments, as described in
  Section \ref{multiple-alignment-and-dashboard}.
\item
  A multi-reference re-annotation of 500 samples from Sova-RuDevices \citep{sova2019}
  test set from scratch (without seeing the original labeling during the
  annotation). We use it further to study fine-tuning dynamics in
  section \ref{fine-tuning-dynamics}.
\end{enumerate}

We argue that annotating audio samples 2-3 minutes long is enough to
evaluate longform models based on segmenting (VAD-based or uniform),
because this length is enough to evaluate the segmenter quality. With
2-3 minutes per sample, we can collect much mor diverse dataset.
Annotating too long audios is costly, while the speaker and topics
diversity would be low.

\section{MWER annotation syntax vs. normalization} \label{fine-tuning-dynamics}

We conducted exepriments to check if the MWER features
matter in practice. Simply introducing annotation errors at random would
raise absolute WER values, but should not typically affect model
ranking, because all the models are penalized equally. This means that
we can perform a meaningful model comparison even if the dataset is
partly erroneous at random. In this light, are "perfect" test sets
important?

Theoretically, we assume that penalties caused by lacking multivariance
are usually \emph{non-random}. Thus, the dataset annotation style can be
learned during fine-tuning: model B can be generally worse then A, but
better fitted to the dataset annotation style. Thus, single-variant
datasets may show misleading dynamics during fine-tuning even with
normalization, causing an illusion of metric improvement.

In our experiments, we use a human-labeled Sova-RuDevices \citep{sova2019} Russian
dataset, originally crowdsourced from voice messages, and fine-tune
\emph{whisper-medium} and \emph{whisper-large-v3-turbo} \citep{radford2023robust} on the train split. As a result, we obtain a
series of checkoints. We then take a part (500 samples) of the test
split and re-annotate it carefully using multi-reference and wildcard
syntax.

We evaluate each checkpoint on the test part in 4 ways (with condifence
intervals):

\begin{enumerate}[topsep=0pt]
\item
  The original labeling.
\item
  The novel multi-referece relabeling
\item
  The original labeling with text normalization similar to \citep{radford2023robust}:
  removing filler words, standardizing number formatting with Silero
  normalizer \citep{silero2020} and converting anglicisms into Russian. We also try lighter normalization
  without a Silero normalizer.
\end{enumerate}

\begin{figure}[h!]
    \centering
    \begin{minipage}{0.48\textwidth}
        \includegraphics[width=\linewidth]{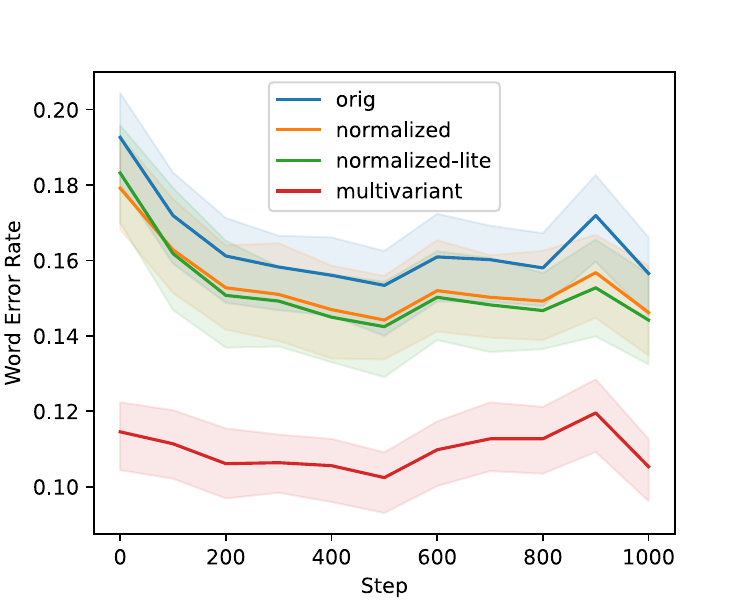}
    \end{minipage}\hfill
    \begin{minipage}{0.48\textwidth}
        \includegraphics[width=\linewidth]{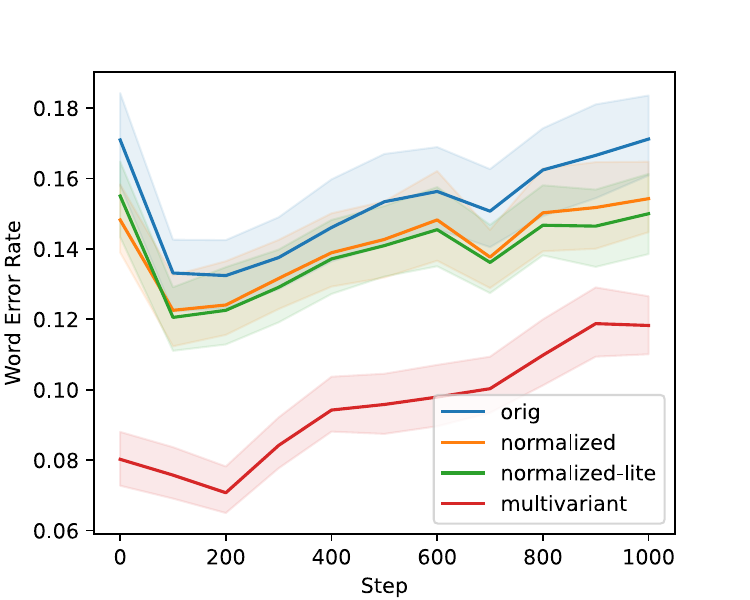}
    \end{minipage}
    \caption{
      \textbf{Upper, left. Does multivariant re-labeling of the test set alter the visible fine-tuning dynamics?}
      We avaluate the same fine-tuning checkpoint series 4 times on the same test set, but with different labeling,
      and observe different dynamics. If we stop after step 800, for \emph{whisper-medium} we get around 3\%
      WER improvement for the original \emph{sova-rudevices} test part with any normalizer, but
      0\% WER improvement for the multivariant one.
    }
    \label{fig:tuning}
\end{figure}

The results are shown in \ref{fig:tuning}. Since the text normalization approach
is apriori approximate, and the multi-reference re-labeling appoach is
more strict, their different dynamics suggest that the model quickly
adopts the dataset labeling style. One of the examples is the Russian
"blyat/blyad" spelling that ocurrs frequently and is spelled
differently in the dataset and by the pretrained model. This shows the
importance of multi-reference labeling to mitigate false metric
improvements.

\section{Conclusion} \label{conclusion}

In this paper we proposed a tweaked multi-reference transcription
alignment algorithm and release a ready to use tool to visualize
multiple alignments, and \textit{asr\_eval} tool for streaming and offline ASR
evaliation and dataset correction. We demonstrated a case study that compares (i)
multi-reference labeling with the wilcard symbol and (ii) text
normalization, and show the benefits of the first. We argue that this
may be important in several cases:

\textbf{In noisy, fluent speech} (such as Sova-RuDevices dataset) the
percentage of multiple options or unclear speech is large. In our
relabeling, around 50\% of 2-5 sec long samples contain multi-reference blocks
or the wilcard symbol.

\textbf{In high-quality clear speech} the percentage of unclear speech
is much lower, but some words still may be written differently. There
are roughly 1-2\% of these words. As the speech recognition systems
become better, their WER may appoach similar numbers. If a model gains
3\% WER while 1.5\% WER are caused by lacking multivariance in
annotation, then half of the model errors are spurious. This highlights
the importance of multivariance even in clear speech.

In general, there are many works from different research areas that show
that proper testing may challenge a large layer of the previously
proposed methods. This happened in image classification \citep{gulrajani2020search},
object detection \citep{lee2022rethinking}, tabular machine
learning \citep{shwartz2022tabular}, NLP and more. Thus, a
large part of machine learning progress is based on robust benchmarking
and ready to use code tools that we tried to contribute in this work.

\bibliography{submission}
\bibliographystyle{fancypaper}

\newpage
\onecolumn
\appendix
\section{The alignment algorithm} \label{app:appendix_algorithm}

Our algorithm is a generalization of the Needleman--Wunsch algorithm to
support multi-choice blocks and wildcard blocks (matching any sequence,
possibly empty) for both sequences.

\textbf{Flat view.} We start by converting a multivariant annotation
into a ``flat view''. Our example contains 9 tokens (words) in total,
including those inside and outside multivariant blocks. We arrange them
into a list, prepend and append position. Then we define transitions
between them, according to the multivariant structure. In our example,
from the ``'' word we can transit to ``uh'' or to ``please'' (skipping
the \{uh\} optional block). The transitions are denoted as arrows on the
vertical axis (for truth) and on the horizontal axis (for prediction).

\begin{figure}[h!]
    \centering
    \includegraphics[width=\linewidth]{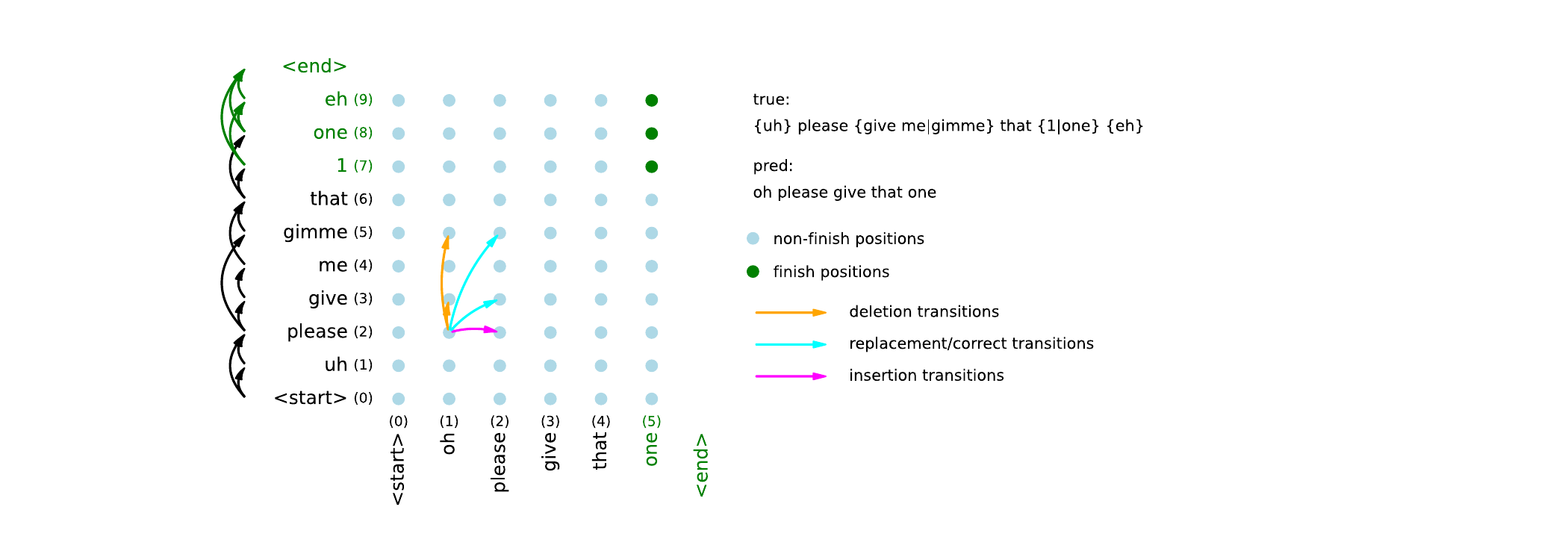}
    \caption{\textbf{Solving the optimal alignment in multi-reference case.}}
    \label{fig:solving_alignment}
\end{figure}

\textbf{Matrix position.} A position in a resulting matrix means some
pair of locations. More precisely, the matrix position ``at token A in
the annotation and token B in the prediction'' means that these tokens
were the last tokens we visited, that is, we are immediately after these
tokens (Figure \ref{fig:solving_alignment}).

\textbf{Matrix transition (step).} A ``step'' is a transition between
matrix positions. There are 3 types of transitions:

\begin{enumerate}[topsep=0pt]
\item
  Vertical: transition over the annotation only. This means a
  ``deletion'' in terms of sequence alignment, because we go to the next
  token at the annotation while staying on the same position at the
  prediction.
\item
  Horizontal: transition over the prediction only. Similarly, this means
  an ``insertion''.
\item
  Diagonal: transition over both the prediction and the annotation. This
  means either ``correct'' or ``replacement'', depending on whether the
  tokens match or not.
\end{enumerate}

\textbf{Step score.} Each step is associated with a score 0 or 1,
depending on how much did the total number of errors increase after the
step. A diagonal step may have a score 0 or 1, depending on whether the
tokens match or not. The horizontal step is always 1, except cases where
the current annotation token is \emph{wildcard} - the score is 0 in this case,
because we can step over any count of prediction tokens being at wilcard
annotation token without incrementing error counter. The same for
vertical step (however, the wildcard tokens in the prediction normally
should not occur).

That is, we can traverse the matrix from the left lower to the right
upper corner, or vice versa, and thus build an alignment, resulting in a
final score (number of word errors) where we finish traversing both
sequences.

\textbf{Position score and best path.} We define the best path for
position as the lowest score path from the end to the current position.
This is the same as the best alignment between the remaining parts of
both sequences (from the current position to the end). The position
score is a score of its best path. Our final goal is to find the best
path for the \verb|(<start>, <start>)| matrix position - this will be the optiomal sequence
alignment.

\textbf{Final positions.} To start with, we define as ``final'' all the
annotation positions leading directly to the \verb|(<end>)| position, and the same for
the prediction positions. In our example, 1 prediction position and 3
annotation positions are final (green), and the corresponding steps to
the \verb|(<end>)| are shown in green. We define a ``final matrix position'' as a
combination of final annotation position and final prediction position
(green dots). Being in a final martrix position, we are after final
tokens in both sequences, that is essentially at the end. So, best path
are empty for the final matrix positions, and their scores are zero.

Now we are going to fill the matrix from the final positions to the \verb|(<start>, <start>)|
position.

\textbf{Matrix filling.} Consider some matrix position X with unknown
score and best path. Let $S$ be a set of all matrix positions where we can
get to in one step from $X$. In our example, the position \verb|X=(please, oh)|
contains 5 positions in $S(X)$: 2 for vertical steps, 2 for
diagonal and 1 for horizontal. Let's assume that we know a position
score for all positions in $S(X)$. For each $Xs \in S(X)$, we can calculate
$step\_score(X, Xs) + score(Xs)$, and take the argmin across $S(X)$ to find
the best source position $X^*$. Thus, $score(X) = step\_score(X, X^*) +
score(X^*)$, and the best path for $X$ consists of one step to $X$ plus
the best path for $X^*$.

Using the described process, we fill the whole matrix, starting from the
last row from right to left, then the next row, and so on.
Alternatively, we can fill the matrix column-wise or even diagonal-wise
(which is the best to parallelze computations). Finally, we got a score
for the position \verb|(<start>, <start>)|, which is the total number of word errors. To
restore the best path (the alignment), it's enough to keep the source
position for each matrix position, and restore the path iteratively.

\textbf{Complexity.} Let $A$ be the total word count in the annotation,
and $B$ be the total word count in the prediction. Also, let $S(X)$ be
bounded above (which is always the case in real multi-reference
annotations). The complexity is $O(AB)$ for the matrix propagation, since
the number of operations to fill a cell is bounded above. The complexity
of restoring the best path is $O(max(A, B))$ that is negligible comparing
to $O(AB)$. Thus, the overall complexity is $O(AB)$. Since transcriptions $B$
usually have length similar to the annotation $A$, the complexity is
quadratic, as in the Needleman--Wunsch algorithm.

\textbf{Improved scoring.} There may be many cases where the argmin of
$step\_score(X, Xs) + score(Xs)$ contains more than one source positions.
This reflects the fact that there may be many optimal alignments. Across
them all, we want to select those in which:

\begin{enumerate}[topsep=0pt]
\item
  The number of correct matches is the highest.
\item
  The mismatched words have the least number of discrepancies in
  symbols.
\end{enumerate}

This is straightforward to implement. Initially, we have a single score:

$n\_errors = n\_replacements + n\_deletions + n\_insertions$

We replace $n\_errors$ with a tuple of three scores:

\begin{enumerate}[topsep=0pt]
\item
  $n\_errors$
\item
  The number of correct matches in the path
\item
  Sum character errors for all the transitions int the path
\end{enumerate}

These tuples are compared lexicographically: to compare $(A1, B1, C1)$
with $(A2, B2, C2)$ we first compare $A1$ with $A2$, if equal we compare $B1$
and $B2$, and if equal we finally compare $C1$ and $C2$. This ensures that the
resulting best path is still optimal in terms of $n\_errors$.

One possible alternative is to align on a character level (calculating
CER instead of WER). Our algorithm is applicable also for this case.
Note that such an alignment will not always be optimal in WER sense.
This is stll a valid alternative, however, CER have its own biases (for
example, it underestimates errors in word endings, since it's only one
or two characters).

\section{The extended table of use cases} \label{app:extended_table}

\begin{xltabular}{\textwidth}{ 
    >{\hsize=0.5\hsize}L 
    >{\hsize=1.0\hsize}L 
    >{\hsize=1.5\hsize}L 
    >{\hsize=1.0\hsize}L 
}
    \caption{\textbf{Multi-reference and wildcard syntax use cases.} We use syntax \texttt{\{A|B|C\}} to enumerate multiple equally acceptable options, \texttt{\{A\}} for optional text blocks, \texttt{\{A|~B\}} for the case where B is acceptable but contains minor typical spelling errors. \texttt{<*>} is a wildcard symbol used where we cannot reliably enumerate all possible choices. Our MWER algorithm can use the provided syntax directly to evaluate WER/CER, aligning \texttt{<*>} with arbitrary sequence without penalty. See also Table 2 for complex cases.}
    \label{use-cases} \\
    \toprule
    \textbf{Use case} & \textbf{Examples} & \textbf{Guideline and commentary} & \textbf{Can text normalization handle this?} \\
    \midrule
    \endfirsthead

    \multicolumn{4}{c}{{\bfseries \tablename\ \thetable{} -- continued from previous page}} \\
    \toprule
    \textbf{Use case} & \textbf{Examples} & \textbf{Guideline and commentary} & \textbf{Can text normalization handle this?} \\
    \midrule
    \endhead

    \midrule
    \multicolumn{4}{r}{{Continued on next page}} \\
    \endfoot

    \bottomrule
    \endlastfoot

    Numerals, currency and units of measurement
    & 
    En: \rustt{\{Fourth|4|4th|4-th\}} \newline En: \rustt{\{10|ten\} \{mm|millimeters\}} \newline Ru: \rustt{\{Первый|1|1-й\}} \newline Ru: \rustt{\{1\%|1 \%|1 процент|один процент\}}
    & 
    Annotate all possible options. Take into account the tokenization scheme that will be used: if space is a separator, and "\%" sign is not removed together as punctuation, then "1\%" and "1 \%" are different options. See also Table 2 for complex cases.
    & 
    Yes, but may be prone to errors, especially for languages where good normalization rules (or model) do not exist yet.
    \\
    \midrule
    Poorly heard inflections
    & 
    En: \rustt{the player's own \{fantasy|fantasies\}}. \newline Ru: \rustt{частичку \{фантазий|фантазии\} игрока}.
    & 
    Enumerate one or multiple options. It depends on the annotator's confidence in what they heard, taking into account the context of the phrase.
    & 
    Typically no.
    \\
    \midrule
    Wrongly spelled inflections
    & 
    Ru: \rustt{Вот наши \{шампура|шампуры\}}.
    & 
    Annotate both options: correct one and the actually spelled one. We justify this by the fact that annotator may decide to fix the spelled inflection or not, both options seem valid.
    & 
    Typically no.
    \\
    \midrule
    Repetitions
    & 
    En: \rustt{And \{the first we need...\} the first we need is a plan.} \newline Ru: \rustt{И \{первое, что нам нужно\} первое, что нам нужно это план.}
    & 
    Annotate as optional, since some transcriber systems remove repetitions, some don't.
    & 
    Typically no.
    \\
    \midrule
    Transliterations, possibly inflected ones
    & 
    Ru: \rustt{в \{Facebook|Facebook-е|Фейсбуке\}}
    & 
    Enumerate all possible options. Sometimes, in case of phonetic ambiguity in the absence of established transliteration, the number of possible options can be high, see Table 2 for complex cases. If transliteration looks inappropriate, keep only the original spelling (example: "Hetman Software" in Russian speech).
    & 
    Sometimes, depending on the rules and models.
    \\
    \midrule
    Poorly heard speech, hesitations.
    & 
    \rustt{<*>}
    & 
    Noisy, fluent speech from in-the-wild sources is a typical use case, but there are many moments that include poorly heard or overlapped speech, interruptions and hesitations. We propose to annotate these places with a wildcard \rustt{<*>} symbol. This is also useful when we cannotate all possible options in complex cases, see Table 2.
    & 
    No.
    \\
    \midrule
    Filler words
    & 
    En: \rustt{\{well\}} \newline Ru: \rustt{\{угу|ага|\} , конечно.} \newline Ru: \rustt{\{Ну\} не совсем.}
    & 
    Annotate as optional. Do not annotate vocalizations (see table 2).
    & 
    The normalization algorithm can remove all the filler words it knows from both the transcription and the annotation. But this makes us unable to penalize incorrect filler word insertions by the transcriber, or penalize cases when the transcriber wrongly removes those words used meaningfully.
    \\
    \midrule
    Minor syntax errors
    & 
    Ru: \rustt{Энергии \{собраны|~собранны\}, \{сконцентрированы|~сконцентрированны\}.} \newline Ru: \rustt{\{неудивительно|~не удивительно\}}
    & 
    Some spelling rules are too complex for many people, even those who are quite literate. We advocate for permissive annotation in these cases, because models trained on human transscriptions will do the same mistakes. In \textit{asr\_eval}, the special label "~" is parsed by the annotator to enable eveluation in a lexically strict or permissive mode. \footnote{This feature is not implemented yet, currently evaluation is always permissive in \textit{asr\_eval}.}
    & 
    Typically no.
    \\
\end{xltabular}

\section{Table of complex cases} \label{app:complex_cases}

\begin{xltabular}{\textwidth}{ 
    >{\hsize=0.5\hsize}L 
    >{\hsize=0.7\hsize}L 
    >{\hsize=1.4\hsize}L 
    >{\hsize=1.4\hsize}L 
}
    \caption{\textbf{Complex cases to annotate.}}
    \label{complex-cases} \\
    \toprule
    \textbf{Case} & \textbf{Example} & \textbf{Problem} & \textbf{Solutions} \\
    \midrule
    \endfirsthead

    \multicolumn{4}{c}{{\bfseries \tablename\ \thetable{} -- continued from previous page}} \\
    \toprule
    \textbf{Case} & \textbf{Example} & \textbf{Problem} & \textbf{Solutions} \\
    \midrule
    \endhead

    \midrule
    \multicolumn{4}{r}{{Continued on next page}} \\
    \endfoot

    \bottomrule
    \endlastfoot

    Rare proper names or terms
    & 
    En: \rustt{Arshankrut Kitra Katjvanmaninik}, \newline \rustt{Ru: Аршанкрут Китора Катьванманиник}
    & 
    Even a perfect transcriber may be not familiar with proper names from specific corpora. The human annotator may also make mistakes.
    & 
    Annotate with the correct option only, if human annotator knows it. This may seem contrary to our rule to include options with minor spelling errors, however, in the case of proper names, misspelling proper names creates a much worse experience for the reader. If the annotator does not know the correct spelling, either use a wildcard symbol, use a dashboard with baseline model predictions to check that models do not write this word in another way that also seems appropriate. Failing to do this may occasionally result in a biased dataset (see section 7).
    \\
    \midrule
    Long numbers
    & 
    66-77-66-77
    & 
    Can be transcribed as 6677-6677 or 66776677 etc. We are limited by the tokenization rule which splits by space or by dash, so "66 77" and "6677" are different options.
    & 
    Enumerate only some options or use a wildcard symbol \rustt{<*>}. In future work, a specific engineering is required.
    \\
    \midrule
    Transliterations
    & 
    "PlayStation" in Russian speech
    & 
    In Russian this can be written as \rustt{"Пл\{е|э\}й\{ \}ст\{е/э\}йш\{е|э\}н"}, where \rustt{\{е|э\}} are phonetically similar, and the space between words is optional. However, out current implementation does not support optional spaces or multivariance inside words.
    & 
    Enumerate only some options or use a wildcard symbol \rustt{<*>}. In future work, a specific engineering is required.
    \\
    \midrule
    Vocalizations
    & 
    Aaa, mmm
    & 
    It looks weird to enumerate all possible spellings, and complicates the annotator's task.
    & 
    Skip vocalizations, becase model typically do not transcribe them (and should not). This however may lead to a problem when a model is generally good, but receives too much penalty for transcribing vocalizations (this is the case with GigaAM v3) which ideally should be penalized less. This can be cured on the transcriber's side by applying postprocessing.
    \\
    \midrule
    Music and background noises
    & 
    & 
    Some models tend to describe music and background noises, especially if there is no speech in the fragment. This looks reasonable.
    & 
    MWER syntax cannot help here, so this is up to practioners.
    \\
    \midrule
    Currency
    & 
    \rustt{\{тысячу рублей|1000 рублей|.....|1,000₽\}}
    & 
    Some multi-reference blocks are too complex.
    & 
    Extend syntax with macros in future work.
    \\
\end{xltabular}

\section{The dashboard interface} \label{app:dashboard}

\begin{figure}[h!]
    \centering
    \includegraphics[width=\linewidth]{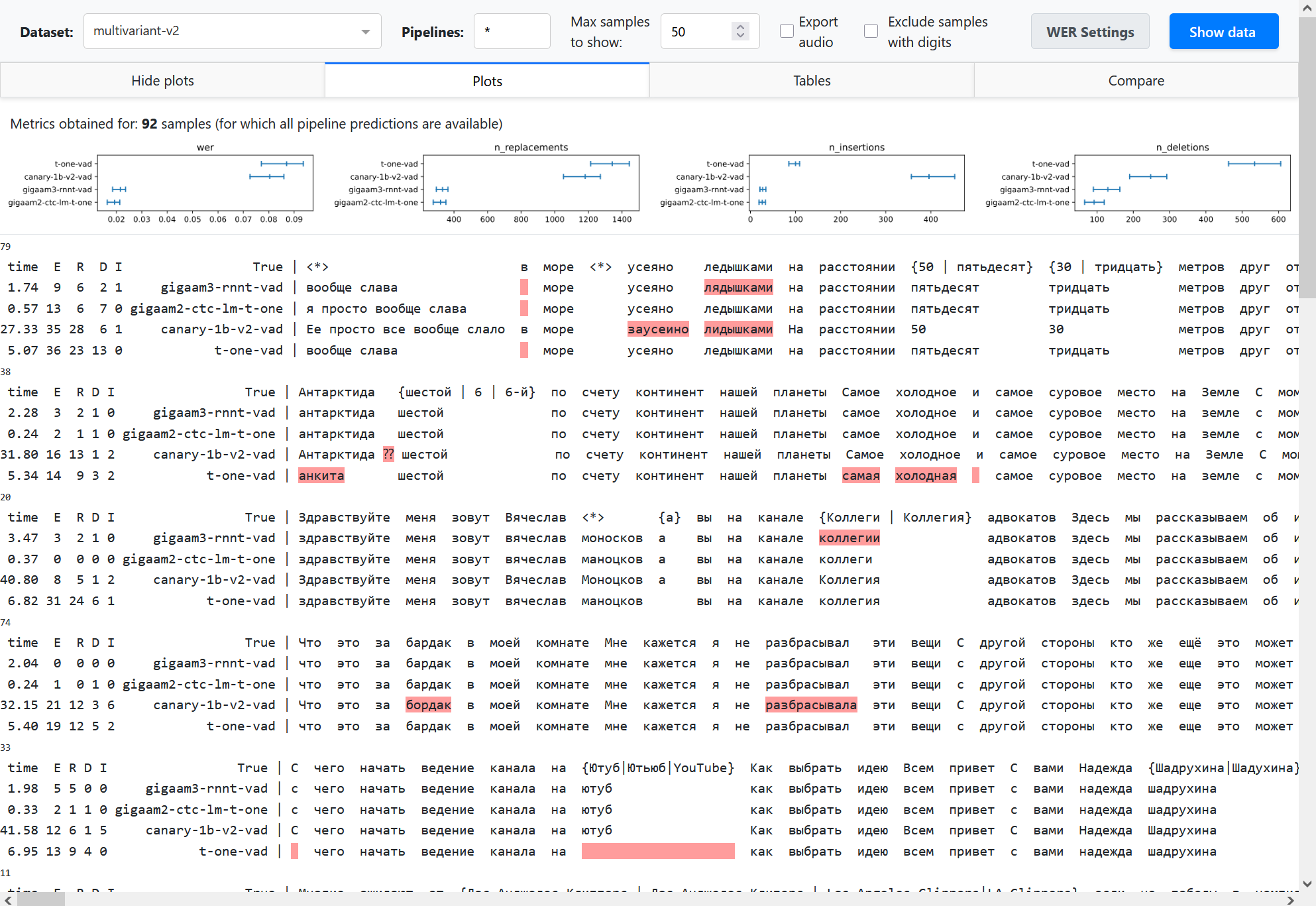}
    \caption{A dashboard screenshot.}
    \label{fig:dashboard_screenshot}
\end{figure}

\end{document}